\documentclass{article}

\usepackage{arxiv}

\usepackage[utf8]{inputenc} 
\usepackage[T1]{fontenc}    
\usepackage{hyperref}       
\usepackage{url}            
\usepackage{booktabs}       
\usepackage{amsfonts}       
\usepackage{nicefrac}       
\usepackage{microtype}      
\usepackage{amsmath}
\usepackage{cleveref}       
\usepackage{lipsum}         
\usepackage{graphicx}
\usepackage{doi}

\usepackage{subcaption}
\usepackage{algorithm}
\usepackage{algpseudocode}
\makeatletter
\algrenewcommand\ALG@beginalgorithmic{\ttfamily}
\makeatother
\usepackage{wrapfig}
\usepackage{setspace}

\usepackage[backend=biber, sorting=none, style=numeric]{biblatex}
\title{Computer-Aided Design of Rational Motions for 4R and 6R Spatial Mechanism Synthesis}

\newif\ifuniqueAffiliation

\ifuniqueAffiliation 
\author{ \href{https://orcid.org/0000-0002-7398-7825}{\includegraphics[scale=0.06]{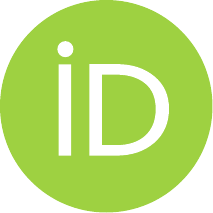}\hspace{1mm}Daniel Huczala}\\
	Unit of Geometry and Surveying\\
	University of Innsbruck\\
	Innsbruck, Austria \\
	\texttt{daniel.huczala@uibk.ac.at} \\
	\And 
	\href{https://orcid.org/0000-0000-0000-0000}{\includegraphics[scale=0.06]{orcid.pdf}\hspace{1mm}Johannes Siegele} \\
	Johann Radon Institute for Computational \\ and Applied Mathematics (RICAM) \\
	Austrian Academy of Sciences \\
	Linz, Austria \\
	\And 
	\href{https://orcid.org/0000-0000-0000-0000}{\includegraphics[scale=0.06]{orcid.pdf}\hspace{1mm}Daren A. Thimm} \\
	Unit of Geometry and Surveying\\
	University of Innsbruck\\
	Innsbruck, Austria \\
        \And 
	\href{https://orcid.org/0000-0000-0000-0000}{\includegraphics[scale=0.06]{orcid.pdf}\hspace{1mm}Martin Pfurner} \\
	Unit of Geometry and Surveying\\
	University of Innsbruck\\
	Innsbruck, Austria \\
        \And 
	\href{https://orcid.org/0000-0000-0000-0000}{\includegraphics[scale=0.06]{orcid.pdf}\hspace{1mm}Hans-Peter Schr\"ocker} \\
	Unit of Geometry and Surveying\\
	University of Innsbruck\\
	Innsbruck, Austria \\
}
\else
\usepackage{authblk}

\newbox{\orcid}\sbox{\orcid}{\includegraphics[scale=0.06]{orcid.pdf}} 
\author[1]{%
	\href{https://orcid.org/0000-0002-7398-7825}{\usebox{\orcid}\hspace{1mm}Daniel Huczala\thanks{\texttt{daniel.huczala@snu.ac.kr}}}%
}
\author[2]{%
	\href{https://orcid.org/0000-0001-9015-1942}{\usebox{\orcid}\hspace{1mm}Severinas Zube}%
}
\author[3]{%
	\href{https://orcid.org/0000-0003-1988-2202}{\usebox{\orcid}\hspace{1mm}Martin Pfurner}%
}
\author[4]{%
	\href{https://orcid.org/0000-0002-8790-0081}{\usebox{\orcid}\hspace{1mm}Johannes Siegele}%
}
\author[1]{%
	\href{https://orcid.org/0000-0002-0293-6975}{\usebox{\orcid}\hspace{1mm}Frank C. Park}%
}
\affil[1]{Robotics Laboratory, Seoul National University, Seoul, Korea}
\affil[2]{Institute of Computer Science, Department of Mathematics and Informatics, Vilnius University, Vilnius, Lithuania}
\affil[3]{Unit of Geometry and Surveying, University of Innsbruck, Innsbruck, Austria}
\affil[4]{J. Radon Instit. for Computational and Applied Mathematics (RICAM), Austrian Academy of Sciences, Linz, Austria}
\fi

\renewcommand{\shorttitle}{CAD of Rational Motions for Spatial Mechanism Synthesis}

\hypersetup{
pdftitle={A template for the arxiv style},
}

\begin{document}

\maketitle

\begin{abstract}
This paper focuses on geometric methods for generating rational motions used in the design of single-loop rational linkages, 1-degree-of-freedom mechanisms that can execute prescribed spatial tasks. Building on established rational motion synthesis methods, we introduce a new interpolation scheme for seven 3D points based on cubic quaternionic Bézier curves. The resulting motion admits factorization, i.e. the synthesis of a spatial six-bar mechanism whose tool frame passes the specified seven points. To support engineering practice, we provide open-source CAD tools that implement also the other methods and provide fast visual evaluation of motion generation and mechanism synthesis. 
\end{abstract}


\newcommand{\qi}{\mathbf{i}}
\newcommand{\qj}{\mathbf{j}}
\newcommand{\qk}{\mathbf{k}}
\def\br{\mathbb{R}} 
\def\bh{\mathbb{H}} 
\def\Im{\mathbf{ Im }} 
\def\Re{\mathbf{ Re }} 

\section{Introduction}\label{sec:intro}

In robot kinematics, the single-loop one degree of freedom (1-DoF) mechanism synthesis problem for a prescribed set of points or poses, known as path generation or motion generation, has attracted interest for decades. 
The objective is to find a motion that would guide the origin of the moving frame through the given input set so that it also allows for the synthesis of a mechanism able to perform the motion. 
In case of planar four-bar mechanisms, it is also known as the Burmester Problem and was solved in multiple ways, for example with five two-dimensional input poses \cite{Hayes2002}. The planar problem is closely related to the spherical case, and the results presented share similarities \cite{Mullineux2011atlasof}. The 1-DoF property provides straightforward and energy-efficient design \cite{Postulka2025}, but its \emph{spatial} applications are more limited than those in the planar or spherical cases.
One of the reasons is that in spatial kinematics the generation of 1-DoF motion is much more challenging. The mechanisms are often overcostrained \cite{Mueller2009overconstr} and there is no meaningful distance metric in three-dimensional special Euclidean group SE(3) \cite{Park1995distanc}, which complicates the numerical and optimization techniques. Therefore, researchers have been investigating algebraic ways for motion generation, and this study also utilizes these findings. 

As is known, there exists only one spatial closed-chain four-bar linkage class (the Bennett mechanism), and there are algebraic methods that construct the Bennett motion that can achieve three given poses \cite{Brunnthaler2005bennett} or five points \cite{Zube2018interpolation}. The resulting motion is parameterized by a rational function and the parameter corresponds directly to the driving joint angle \cite{Huczala2025direct}. The design parameters of the mechanism can then be obtained using the theory of \emph{rational motion factorization} \cite{Hegedus2013factorization}, which decomposes the motion curve into linear factors that represent rotations (materialized by revolute joints). This powerful algebraic algorithm can also be applied to motions of higher degree, as applied in \cite{Hegedus2015fourpose} for four arbitrary poses in space interpolated by a cubic motion, which corresponds to 1-DoF six-bar mechanisms. 
In this paper, we extend the motion generation methodology of \cite{Zube2018interpolation}, provide updated explicit formulas for the case of interpolation through five points, and introduce the interpolation scheme for seven points in a similar manner. The output motions are returned in a factorizable form, i.e. the four- and six-bar mechanism synthesis can be performed immediately.
Additionally, an open-source implementation is provided not only for these new interpolation methods, but also for the algorithms of \cite{Brunnthaler2005bennett}\cite{Hegedus2015fourpose}, including mechanism synthesis using the factorization method, all as an interactive CAD tool (see video in supplementary material \cite{motion_desing_zenodo}). 

The paper is organized as follows: Sect. \ref{sec:math} provides mathematical definitions, Sect. \ref{sec:methods} briefly summarizes the known methodologies for interpolating three and four poses and then guides the reader through the problematic of interpolating five and seven points. Finally, details on the implementation are provided, along with information on supplementary material.

\section{Mathematical Preliminaries}\label{sec:math}



To represent proper rigid body motions of the special Euclidean group $\mathrm{SE}(3)$, we use dual quaternions 
$ \mathbb{DH} =\{p+\varepsilon q\ | \    p,q \in \mathbb{H}, \varepsilon^2=0\} $, alternatively known as Study parameters. They extend regular quaternions $\mathbb{H} \cong \mathbb{R}^4$ in the following way. Given quaternions $p, q$, a dual quaternion $c$ can be written as
\begin{equation}
        c = p + \varepsilon q
        = p_0 + p_1 \qi + p_2 \qj + p_3 \qk + \varepsilon (q_0 + q_1 \qi + q_2 \qj + q_3 \qk), 
\end{equation}
where $\varepsilon$ is the dual unit that squares to zero. The conjugation of dual quaternions is defined as $c^* := p^* + \varepsilon q^*$  and the $\varepsilon$-conjugation is defined as $c_\varepsilon := p - \varepsilon q$. 

The elements of $\mathbb{DH}$ must lie in the \emph{Study quadric} $S\subset \mathbb{DH}$ 
\begin{equation}
 S=\{ p+\varepsilon q\ |\  p q^* + q p^* =2(p_0q_0+p_1q_1+p_2q_2+p_3q_3)= 0\} 
 \label{studyquadric}
\end{equation}
 to represent a rigid body displacement. For more details, including the exceptional space on $S$ 
 or dual quaternion multiplication rules, see \cite[Chap.~13]{bottema1979}.

We will identify 3D points $x=(x_1, x_2, x_3) \in \mathbb{R}^3$ with imaginary quaternions $\mathbf{x}=x_1 \qi + x_2 \qj + x_3 \qk\in\Im(\bh)$\footnote{For some conventions, e.g. \cite{Huczala2025direct}, $\mathbf{x}=-\frac{1}{2}(x_1 \qi + x_2 \qj + x_3 \qk)$ is required.}. Using this identification, we associate a point $p+\varepsilon q$ on the Study  quadric with a 3D point using the map $\pi$
\begin{equation}
    \pi(p+\varepsilon q)= qp^*/pp^*= qp^{-1},
    \label{eq:pi}
\end{equation}
where $p^{-1}=p^*/pp^*$ is an inverse quaternion. The action of a dual quaternion $c=p+\varepsilon q$ on a point $x$ is then
\begin{equation}
    x \mapsto  p\mathbf{x}p^{-1}+qp^{-1}.
\end{equation}
Note that the first summand corresponds to the rotational part of the displacement, while the second summand describes the translational part. Therefore, \eqref{eq:pi} maps a dual quaternion to the image of the origin under this displacement, i.e., the map  $\pi$ extracts the coordinates of the origin of the displaced frame.

A one-parametric motion corresponds to a curve of a form $C(t) \in S$ and can be written in the form of a dual quaternion polynomial $C=p+\varepsilon q \in \mathbb{DH}[t]$. It describes a rigid body motion if and only if $CC^* \in \mathbb{R}[t] \setminus \{0\}$, i.e. it lies on the Study quadric ($C(t)\in S$) or equivalently, $qp^*,\, qp^{-1}\in\Im(\bh[t])$. The trajectories of points in these motions follow rational trajectories parameterized by $t$.
As mentioned above, $C(t)$ can be factorized \cite{Hegedus2013factorization} to synthesize mechanisms. 

\section{Methods to Construct Rational Motions}\label{sec:methods}

\subsection{Quadratic Motion through 3 Poses} 

Three poses $c_{0..2}$ define a plane in the space of dual quaternions uniquely. This plane intersects $S$ in \eqref{studyquadric} in a \emph{conic}, which is the quadratic motion curve to be constructed. Such curves were classified and studied extensively in \cite{hamann2011linesymmetric}. According to \cite{Brunnthaler2005bennett} the conics passing through $c_{0..2}$ can be parameterized in the form
\begin{equation}
    C(t) = \alpha c_0 + (c_1 - \alpha c_0 - \beta c_2) t + \beta c_2 t^2,
    \label{eq:quadratic}
\end{equation}
where $\alpha$ and $\beta$ must be chosen so that the entire curve lies on $S$. Substitution of $C(t)$ into $S$ produces a quartic equation $M(t)$ in $t$. Due to the parameterization used in \eqref{eq:quadratic} the coefficients of $t^0$ and $t^4$ in M(t) vanish and it factorizes into $M(t) = t(t-1)G(t)$ with a linear polynomial $G(t)$. This must vanish independent of $t$ and, therefore, its coefficients yield a linear system of equations for $\alpha$ and $\beta$.

\subsection{Cubic Motion through 4 Poses}

For interpolating four poses with a cubic motion, we give a short summary of the algorithm presented in \cite{Hegedus2015fourpose}.
Four poses span a three-space in the dual quaternions which intersects $S$ in a quadric containing all four poses $c_{0..3}$. 
An interpolating cubic motion exists if this quadric contains straight lines.
Assuming such lines exist, the two straight lines through $c_0=1$ must contain two vectorial dual quaternions $k_1$ and $k_2$. Any twisted cubic on the given quadric must intersect one of these two lines in two points. Thus, let us assume that it intersects the line $t-k_1$ twice and define $c_4=\lambda- k_1$ for some $\lambda\in\mathbb{R}$. The parameter values for which the cubic interpolates $c_0,\ldots,c_4$ are given by the parameters at which the rulings through the given poses intersect the line parametrized by $t-k_1$. From this condition, we can compute $t_0=\infty$, $t_1,\ldots , t_4=\lambda$. With these parameter values, we can compute the Lagrange polynomial $C$ that interpolates $\lambda_i c_i$ at $t_i$ and indeterminates $\lambda_i$ for $i=0,\ldots, 3$. To also interpolate the last pose $c_4$, we can solve $C(t_4)=\lambda_4 c_4$ for $\lambda_i$. Note that the algorithm could also use $k_2$ as input which gives a \emph{second solution}. Further, the obtained polynomial depends on the chosen $\lambda$, i.e., the algorithm of \cite{Hegedus2015fourpose} results in two sets of 1-parameter families of cubic motions. 
Control of these degrees of freedom is also implemented in the CAD tool presented later.

\subsection{Motion through 2n+1 Points}

In contrast to interpolation through poses, the next sections interpolate via 3D points. The construction via quaternionic Bézier curves is applied, i.e., with control points in $\mathbb{DH}$ lying on $S$. The quaternionic weights are specified in such a way that the resulting rational curve lies on Study quadric $S$ and is a proper continuous motion. 



We choose arbitrary $2n+1$ 3D points  $a_i, \ i=0,1,\ldots,2n$, which are the points \emph{to be interpolated}. Let $n \geq 1$ be an integer, 
$T=[ t_0 , \ldots ,   t_n ]$ 
be $n + 1$ pairwise different real values. Using   
\begin{equation}
f_i (t)  =  \prod_{i\not=k}(t-t_k),\ i=0,1, \ldots ,n,
\end{equation}
we define quaternionic polynomials of degree $n$ and a motion polynomial $C(t)$
\begin{eqnarray}\label{C_t}
p(t)=\sum_{i=0}^n  w_{2i}\, f_i(t),\quad 
q(t)= \sum_{i=0}^n a_{2i}\, w_{2i}\, f_i(t),\quad C(t)=p(t)+\varepsilon q(t),
\end{eqnarray}
where weights $w_0,w_2,..,w_{2n}$ are quaternions which will be determined later. 
First of all, note that
\begin{equation}
\pi(C(t_i)) = a_{2i} \in \Im (\bh) = \br^3,\ i=0,1,...,n,
\end{equation}
for arbitrary weights $w_{2i},i=0,1,...,n$ (because $f_j(t_i)=0, j\not=i$). This shows that $C(t)$ already interpolates the points $a_{2i}$ for $i=0,\ldots,n$. To ensure that also the remaining points are interpolated, we set up a system of equations to determine the appropriate weights $w_i$. For this, let $T_s=[ s_1 , ... ,   s_n ]$ be different real parameters such that $T_s\cap T=\emptyset$ and for which the points $a_{2i+1}$ are interpolated.

We consider the linear system
$\pi(C(s_j))= a_{2j-1},\ j=1, ..., n. $
Explicitly this linear system is equivalent to
\begin{eqnarray}
 \sum_{i=1}^n (a_{2i}-a_{2j-1}) \, w_{2i} \,f_i(s_j)=( a_{2j-1}-a_0)\, w_{0}  \,f_0(s_j),\ j=1,...,n. \label{interpolatory_eq}
\end{eqnarray}
Without loss of generality, we may assume that $w_0=1$. 
The quaternion linear system \eqref{interpolatory_eq} with $n$  equations and $n$ unknowns $w_2,w_4,...,w_{2n}$ has a
unique solution if the corresponding matrix $A$ is not singular. Theorem 1 in \cite{Zube2018interpolation} claims that for a {\it general} choice of points $a_i, i=0,...,2n+1$ there is a unique solution.

It is left to show that the motion obtained $C(t)$ meets the Study condition. For this, we need to show $pq^*+qp^*=0$. This polynomial is of degree $2n$ and is the primal part of $\pi(C(t))$. Hence, it is fulfilled for the $2n+1$ parameter values $t_0,\ldots,t_n$ and $s_1,\ldots s_{n}$. Therefore, we have a polynomial of degree $2n$ which vanishes for $2n+1$ parameters, thus has to be $0$.

In the following sections, we present explicit formulas for cases $n=2,3$, which result in quadratic and cubic motions.

\subsection{Quadratic Motion through 5 Points}
For the quadratic motion, we fix five 3D points (via points)
 $a_0,...,a_4$ and parameter values\footnote{The order matters as $a_0$ will be interpolated at $t = 0$, $a_1$ at $t = \frac{1}{4}$, $a_2$ at $t = \frac{1}{2}$, etc.} (via times)
 $T=[0, \frac{1}{2}, 1]$ and $T_s=[\frac{1}{4}, \frac{3}{4}]$. The linear system defined by \eqref{interpolatory_eq} can be solved (see  \cite{Zube2018interpolation}).
 Weights $w_2,w_4$ are
 \begin{eqnarray}
w_2 &=& (-9\, d_{41}^{-1}\, d_{21}-3\, d_{43}^{-1}\, d_{23})^{-1}\,( 9\, d_{41}^{-1}\, d_{10}- d_{43}^{-1}\, d_{30}) \, w_0, \\
w_4 &=& (\ -\, d_{21}^{-1}\, d_{41}- 3\,d_{23}^{-1}\, d_{43})^{-1} \, (3 \, d_{21}^{-1}\, d_{10}+d_{23}^{-1} \,d_{30})\,w_0, 
\label{w4}
\end{eqnarray}
where $d_{ij}=a_i-a_j$. Choosing $w_0=1$ we compute the motion $C(t)$ using \eqref{C_t}.

To obtain the curve $C(t)$ in Bézier form, we express the interpolation basis $f_{0..2}$ by quadratic Bernstein polynomials $\beta_{i}(t) = {2 \choose i}(1-t)^{2-i}t^i,i=0,1,2$
\begin{equation}\label{fiquadratic}
 f_0=\frac{1}{2} \beta_0- \frac{1}{4} \beta_1,\quad f_1=- \frac{1}{2}  \beta_1,\quad 
f_2=-\frac{1}{4} \beta_1+\frac{1}{2} \beta_2.    
\end{equation} 
We substitute expression (\ref{fiquadratic}) to primal $p$
\begin{eqnarray*}
    p&=&w_0f_0+w_2f_1+w_4f_2 =\\
    &=& \frac{1}{2} w_0 \beta_0+ \left(-\frac{1}{4}w_0-\frac{1}{2}w_2 -\frac{1}{4}w_4 \right) \beta_1 +\frac{1}{2}w_4\beta_2 =\\
    &=& u_0 \beta_0 + u_1 \beta_1 + u_2 \beta_2,
\end{eqnarray*}
and dual part $q$ of the motion $C=p+\varepsilon q$
\begin{eqnarray*}
    q&=&a_0 w_0 f_0 + a_2 w_2 f_1 +a_4w_4f_2 =\\
    &=& a_0 \frac{1}{2} w_0 \beta_0 - \frac{1}{4} \left(a_0 w_0+2a_2 w_2 +a_4 w_4 \right) \beta_1 +\frac{1}{2}a_4w_4\beta_2 =\\
    &=& p_0 u_0 \beta_0 + p_1 u_1 \beta_1 + p_2 u_2 \beta_2.
    \end{eqnarray*}
So we obtain the following weights and points
\begin{eqnarray*}
u_0=\frac{1}{2}w_0,& u_1=-\frac{1}{4}(w_0+2w_2+w_4),\hphantom{------} & u_2=\frac{1}{2}w_4,\\
p_0=a_0\hphantom{-}, & p_1=-\frac{1}{4}(a_0w_0+2a_2w_2+a_4w_4) (u_1)^{-1}, & p_2=a_4.
\end{eqnarray*}
Then the motion $C(t)$ can be presented in the Bézier from 
\begin{equation}
    C(t)= \left(\sum_{i=0}^2 u_i\,\beta_i\right)
     +\varepsilon
    \left(\sum_{i=0}^2 p_i\,u_i\,\beta_i \right)
\end{equation}
with the 8-tuples $u_i+\varepsilon\, p_i\, u_i \in \mathbb{DH}, i = 0, 1, 2$ being the three homogeneous control points of the 8-dimensional quadratic Bézier curve that lies on $S$ and can be factorized to synthesize a mechanism. 

\subsection{Cubic Motion through 7 Points}


For the cubic motion we fix seven 3D points $a_0,...,a_6$ and parameter values 
$T=[0, \frac{1}{3}, \frac{2}{3}, 1]$ and $T_s=[\frac{1}{6}, \frac{1}{2}, \frac{5}{6}]$,
and we will need to determine four control points and their weights. 
The linear system defined by \eqref{interpolatory_eq} can be written as 
\begin{eqnarray}
 l_j:c_{j,2}\, w_2  +  c_{j,4}\, w_4 +  c_{j,6}\, w_6  &=& c_{j,8}\, w_0,\ (j = 1,2,3),\label{l1} 
\end{eqnarray}
%
%
where the coefficients $c_{j,i}$ are given by
 \begin{eqnarray}
\left( \begin{array}{cccc}
   c_{1,2}   &c_{1,4}& c_{1,6} &c_{1,8}  \\
   c_{2,2}   &c_{2,4}& c_{2,6} &c_{2,8}  \\
   c_{3,2}   &c_{3,4}& c_{3,6} &c_{3,8} 
 \end{array}\right)=
\left( \begin{array}{rrrr}
   15\, d_{21}   & 5\, d_{41} & 3\, d_{61} & -15\, d_{10}  \\
   9\, d_{23}    & 9\, d_{43} & -3\,d_{63} & 3\, d_{30}  \\
   -5\, d_{25}   & -15\, d_{45} & 15\, d_{65} & -3\, d_{50}  
 \end{array}\right)\label{cij} .
 \end{eqnarray}
with $d_{ij}=a_i-a_j$. We assume that $w_0=1$ and eliminate the unknown $w_2$ using 
 $c_{1,2}^{-1}\,l_1-c_{2,2}^{-1}\,l_2$ and $c_{1,2}^{-1}\,l_1-c_{3,2}^{-1}\,l_3 $, which can be written as
 \begin{eqnarray}
m_j:    e_{j,4} w_4 +  e_{j,6} w_6 = e_{j,8}, \ (j=2,3)  
\end{eqnarray}
\vskip-0.8cm
 \begin{eqnarray}
     \left( \begin{array}{ccc}
       e_{2,4}    &   e_{2,6} &  e_{2,8} \\
          e_{3,4}    &   e_{3,6} &  e_{3,8} 
     \end{array}\right)=
 \left( \begin{array}{ccc}
       h_{4,2,4}    &   h_{6,2,6} &  h_{8,2,8} \\
       h_{4,3,4}    &   h_{6,3,6} &  h_{8,3,8} 
     \end{array}\right)\label{eij}    
 \end{eqnarray}
where $h_{i,j,k}=E(c_{1,2},c_{1,i},c_{j,2},c_{j,k})$, with $E(a,b,c,d)= a^{-1} b-c^{-1}d$ for arbitrary quaternions $a,b,c,d$. 
Similarly, we eliminate $w_6$ and $w_4$ from equations $m_2$ and $m_3$ by taking  
$e_{2,6}^{-1}\,m_2-e_{3,6}^{-1}\,m_3$ leading to $r_{4,4}\, w_4=r_{4,8}$ and
$e_{2,4}^{-1}\,m_2-e_{3,4}^{-1}\,m_3$ leading to $r_{6,6} \,w_6=r_{6,8}, $ where
\begin{eqnarray}
r_{4,4}=\ E(e_{2,6},e_{2,4},e_{3,6},e_{3,4}),\
r_{4,8}=\ E(e_{2,6},e_{2,8},e_{3,6},e_{3,8}),\ \label{r1} \\
r_{6,6}=\ E(e_{2,4},e_{2,6},e_{3,4},e_{3,6}),\ 
r_{6,8}=\ E(e_{2,4},e_{2,8},e_{3,4},e_{3,8}),\ \label{r2} \\
w_4=\ r_{4,4}^{-1}\, r_{4,8},\  
w_6=\ r_{6,6}^{-1}\, r_{6,8}.\hphantom{-------------}\label{w2w4}  
\end{eqnarray}
Using \eqref{l1} we compute $w_2$ by $w_2= c_{1,8}-c_{1,4} w_4-c_{1,6} w_6.$
Therefore, we found a parametrization of a cubic curve by the formula \eqref{C_t}. The parametrization $\pi(C(t))$ interpolates points
$a_0,...,a_6$ at the parameter values 
$\frac{i}{6},i=0,..,6$.

To present the curve in Bézier form, we need to convert the interpolation basis $[f_0,...,f_3]$ to Bézier basis 
$[\beta_0,...,\beta_3],$ where $ \beta_i(t)={3 \choose i}(1-t)^{3-i}t^i$ as:
\begin{eqnarray}
    f_0 = -\frac{2}{9} \beta_0 + \frac{5}{27} \beta_1-\frac{2}{27} \beta_2,\quad  
f_1=\hphantom{} \frac{2}{9} \beta_1-\frac{1}{9} \beta_2, \\
f_2= \frac{1}{9} \beta_1-\frac{2}{9} \beta_2,\quad  
f_3= \frac{2}{27} \beta_1-\frac{5}{27} \beta_2 +\frac{2}{9} \beta_3. \hphantom{-} 
\end{eqnarray}
Now we may write the interpolated cubic (defined by the formulas \eqref{C_t}) in Bézier form in the same way (after substitution in $p,q$) as in the previous section
\begin{equation}
    C(t)=(u_0 \beta_0 +...+ u_3 \beta_3) +\varepsilon\, (p_0 u_0 \beta_0 +...+ p_3 u_3\beta_3) ,
\end{equation}
where
\begin{eqnarray*}
 u_0= -\frac{2}{9}\,w_0,\\
     u_1=\frac{5}{27}w_0+\frac{2}{9}w_2+\frac{1}{9}w_4+\frac{2}{27}w_6,\\
      u_2=-\frac{2}{27} w_0 -\frac{1}{9} w_2 -\frac{2}{9} w_4-\frac{5}{27}w_6,\\ 
      u_3=\frac{2}{9}w_6,\\
      p_0= a_0,\\
       p_1= \left(\frac{5}{27}a_0 w_0+\frac{2}{9}a_2 w_2+\frac{1}{9}a_4 w_4+\frac{2}{27}a_6 w_6\right)u_1^{-1},\\
      p_2=\left(-\frac{2}{27}a_0 w_0 -\frac{1}{9}a_2 w_2-\frac{2}{9}a_4 w_4-\frac{5}{27}a_6 w_6\right)u_2^{-1},\\
      p_3= a_6.    
\end{eqnarray*}

\subsection{Implementation}

\begin{figure}[tb]
    \centering
    \begin{subfigure}[t]{0.63\textwidth}
        \centering
        \includegraphics[width=\textwidth]{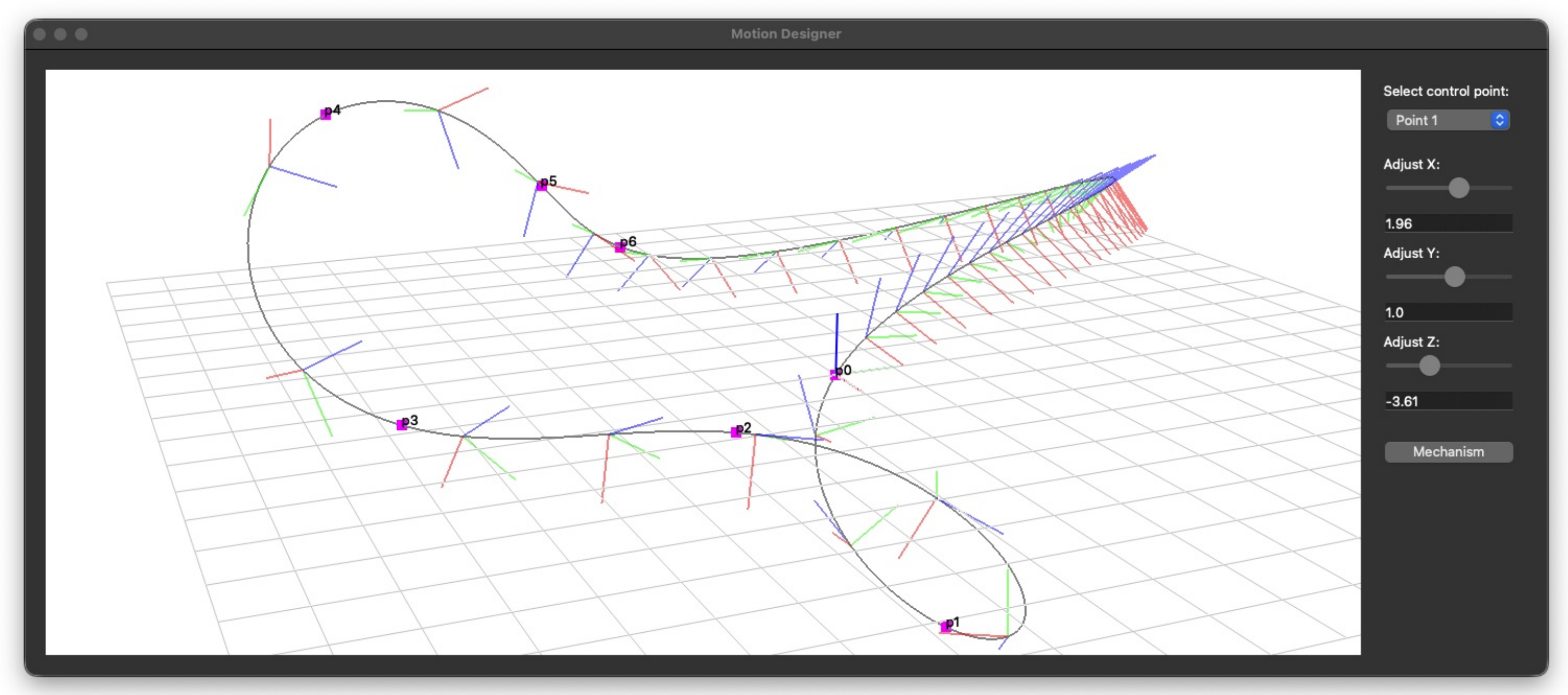}
    \end{subfigure}
    \hfill
    \begin{subfigure}[t]{0.36\textwidth}
        \centering
        \includegraphics[width=\textwidth]{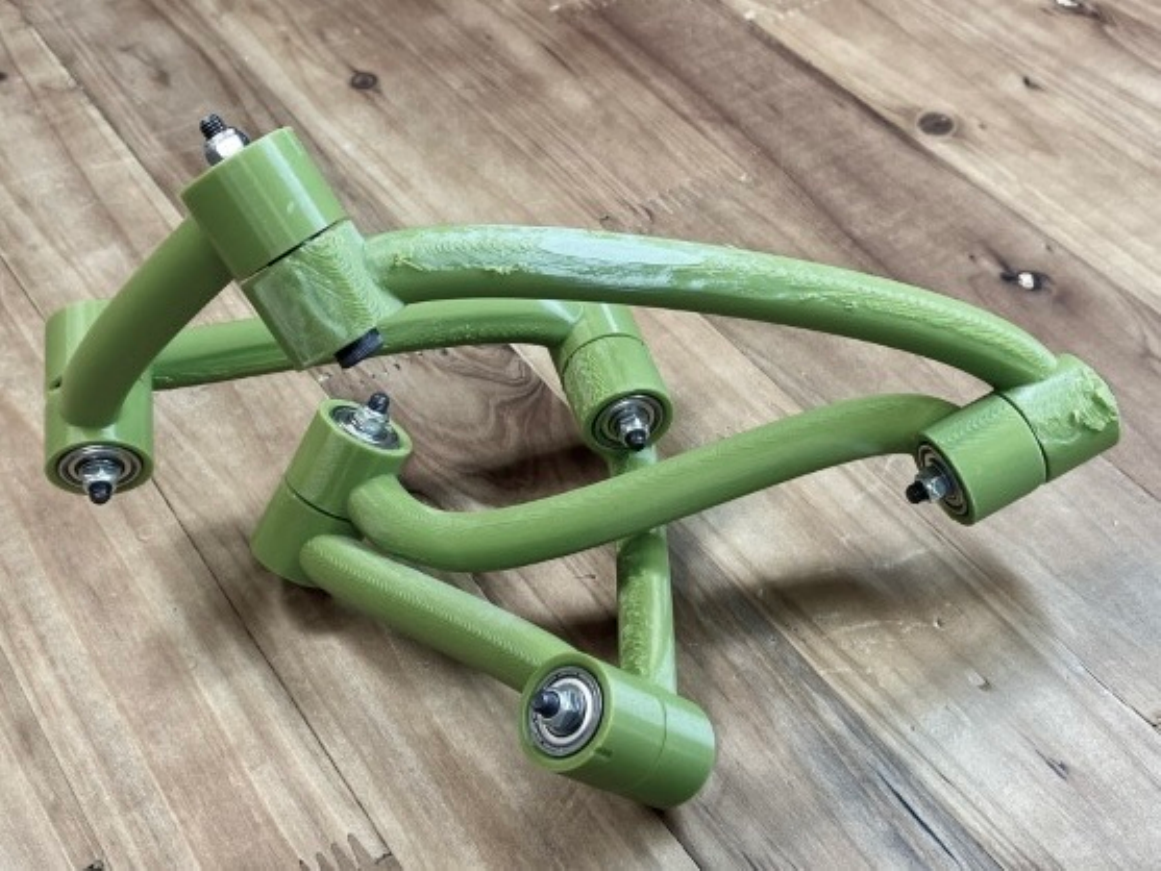}
    \end{subfigure}
    \caption{Left: Motion Designer -- an interactive plot of a 1-DoF motion that interpolates 7 given points (purple); right: example of a produced 6R prototype.}
    \label{fig:motion_des}
\end{figure}

For the kinematics community, we implemented all methodologies as an interactive CAD module called Motion Designer in the Rational Linkages library \cite{huczala2024linkages} version 2.4, an open-source Python package for rapid prototyping of mechanisms. Some crucial calculations were compiled using the Rust language to provide fast rendering with Qt6 software. The visual CAD interface is shown in Fig. \ref{fig:motion_des} with a 6R prototype designed using this methodology. 
For more details and source code investigation, see the supplementary material. \cite{motion_desing_zenodo}.


\section{Conclusion}

This study provided a way to interactively construct rational motions interpolating three or four poses and five or seven points. In addition to state-of-the-art, closed-form solutions for the latter interpolations of points with rational curves in $\mathrm{SE}(3)$ were presented, along with the implementation providing visual CAD software. The output motions serve as the key concept for the synthesis of 1-DoF single-loop mechanisms using rational motion factorization \cite{Hegedus2013factorization}, resulting in spatial four- and six-bar mechanisms.
The algorithms are part of the Rational Linkages library~\cite{huczala2024linkages}, version 2.4 or higher. The interactive use case is shown in the accompanying video together with web-links to the source code, documentation, and tutorials, and can be found in \cite{motion_desing_zenodo}.

\section*{Acknowledgments}

 We gratefully acknowledge the support of the bilateral Austrian–Korean project “Surface generation via integrable evolution of curves,” which is funded by OeAD-GmbH (Austria’s Agency for Education and Internationalization, project no. KR 02/2025), and the National Research Foundation of Korea (project no. RS-2025-1435299).
 This work was also supported by the Vizerektorat für Forschung of the University of Innsbruck.




\printbibliography

\end{document}